# Semi-Supervised Representative Region Texture Extraction of Façade

Zhen Ni, Guitao Cao, Ye Duan

**Abstract.** Researches of analysis and parsing around façades to enrich the 3D feature of façade models by semantic information raised some attention in the community, whose main idea is to generate higher resolution components with similar shapes and textures to increase the overall resolution at the expense of reconstruction accuracy. While this approach works well for components like windows and doors, there is no solution for façade background at present. In this paper, we introduce the concept of representative region texture, which can be used in the above modeling approach by tiling the representative texture around the façade region, and propose a semi-supervised way to do representative region texture extraction from a façade image. Our method does not require any additional labelled data to train as long as the semantic information is given, while a traditional end-to-end model requires plenty of data to increase its performance. Our method can extract texture from any repetitive images, not just façade, which is not capable in an end-to-end model as it relies on the distribution of training set. Clustering with weighted distance is introduced to further increase the robustness to noise or an imprecise segmentation, and make the extracted texture have a higher resolution and more suitable for tiling. We verify our method on various façade images, and the result shows our method has a significant performance improvement compared to only a random crop on façade. We also demonstrate some application scenarios and proposed a façade modeling workflow with the representative region texture, which has a better visual resolution for a regular façade.

## 1. Introduction

Modeling a realistic façade is an important part of the inverse modeling of the building. Traditional 3D reconstruction scheme by aerial oblique images [1,9] can construct an approximate shape of a building, but lacks the detail of the façade. Researches of analysis and parsing around facades [2] to enrich the 3D feature of facades by semantic information raised some attention in the community, like [3] focused on the analysis and refined modeling of window of the façade.

We notice that most façades are regular and repetitive, and the texture of a façade background is also repetitive. Therefore, an assumption arises: we may use an area inside the façade to represent the global texture feature of a façade. This texture is called a representative region texture. Figure 1 shows some examples of this explanation. And in a practical façade modeling task, this representative texture can be tiled over the façade background part instead of directly pasting the original façade image to increase the visual resolution of models (we will discuss more in section 5).

As for the data source of a representative region texture, we select façade image as input format, or more specifically a street view images directly facing the target façade. Recent years, object detection with deep learning has achieved great progress. Studies like [4] uses object detection with deep learning to detect façade components including windows and doors. The task of façade texture extraction is somehow similar to an object detection task: the target object is now a representative texture inside a façade, which is surrounded by the bounding box, using the same network but different data. This method is somehow end-to-end that we use a single network to do the whole task. But this method has a limitation, which is we must construct plenty of data to increase the model performance. Among most public dataset, there doesn't exist one that meet our needs. Mark out all representative texture is obviously too expensive, training from scratch using only a few samples is unstable [12], and transfer learning like using the pretrained weights learned from a larger dataset and fine-tuning on smaller dataset [10,38] for façade-related recognition seems doesn't work well as detailed structure and similar candidate regions may not be recognized by pretrained models [3,11]. Using an end-to-end model also hides its internal details, making it like a black box and lack of interpretability [13]. Moreover, we hope that the implementation of some internal parts of our non-end-to-end model can inspire others or even be reused, such as using representative region extraction on all repetitive images, not only façades.

This paper proposes a semi-supervised way to do representative region texture extraction. All works are based on existing datasets, no additional dataset is constructed. First step is to do semantic segmentation on façade, so we can separate façade background part from the image, and this is the only supervised part. Then on the background part, do lots of random sampling and get lots of regions that may contains the representative texture which we called as candidate textures. Dimension reduction is done on these candidate textures and send outputs to clustering, and we will consider the candidate closest to the center of the maximum cluster is the most representative one. Detailed description will show in Section 3.

In Section 2, we look through some related works. Section 4 shows experimental evaluations and limitations of our method. In Section 5, we will show how to use representative region texture in some practical tasks, such as inverse procedural façade modeling or scene reconstruction with textured voxels.

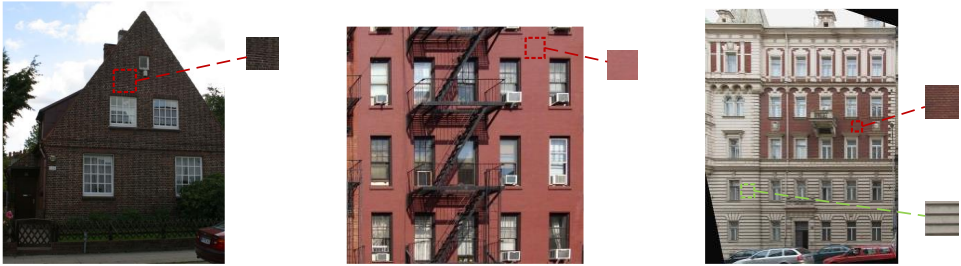

**Figure 1.** Some examples of our task. These three facades are all suitable for extracting a certain region inside of it as the texture that represents the main texture feature of the façade. Left two façades can use one single representative texture to represent the whole façade which is done by our work, while the façade on the right side should use two representative textures to represent different part of the façade, which is not completely resolved in our work (see section 4.3 for more).

## 2. Related works

This section briefly reviews some related topics used in the literature.

**Façade parsing.** Many works have researched the parsing of building façades. Traditional approach focused on utilize the regular arrangements of objects to infer the structure of façade through grammatical rules [16,17,24], which required some parameters of the syntax in advance and cannot cover the diversities in real world. [22] used a Conditional Random Field (CRF) and spatial pattern templates to capture the spatial relations and detect targets. In recent years, deep learning approach such as Convolutional Neural Network [14] have achieved impressive results in computer vision tasks like image classification [15]. As for façade parsing, two typical image processing approach of deep learning are commonly adopted, object detection [23] and semantic segmentation [18].

For object detection, rectangular regions are used to include the target components of façade. Object detection task with CNN is very slow due to it needs to slide the window through the whole image. Some researches focused on first extract the proposal region and then do classification on these candidates, like R-CNN [23,25] did. YOLO [26] merged these two steps into one step, only requires a single integrated forward computation. YOLO v3, an upgraded version of YOLO, integrated ResNet [27] and FPN [28] into its model, the detection accuracy and speed are improved furthermore. [4] used YOLO v3 as its object detection model to do façade component detection.

Semantic segmentation can better segment the boundaries of targets in pixel level. [18] proposed to transform a standard classification used CNN to a fully convolutional network (FCN) for semantic segmentation. UNet [19] add some skip connections in order to preserve both high-resolution detail features and low-resolution rough features. Deeplab [20] uses atrous convolution [21] and atrous spatial pyramid pooling to both enlarge receptive fields and preserve features under different scales. Since the task we do needs a refined segmentation result, semantic segmentation is adopted in our solution.

**Façade and urban modeling.** Mainstream façade and urban modeling process rely on asset extraction. Earlier works on façade modeling generally used shape grammers or low-level features to parse or estimate façade structures and then reconstruct the façade with structure [6,7,56]. CityEngine [57] relies on a reusable and tileable textures and prefab library to make building models more realistic and diverse. [5,8] use lidar and point cloud to detect and reconstruct building façades. Recently deep learning methods are also widely adopted for façade modeling. [3] proposes an adversarial training strategy for the semi-supervised window structure recognition and a 3D façade window modeling approach. [53] collects large amount of urban panoramic imagery and proposes a system for extracting architectural assets from them, including synthesizing façade layouts with GAN and asset detection with Mask R-CNN [58]. We will show a façade modeling workflow using the representative texture in section 5 which is similar to the working process of CityEngine.

**Texture classification.** The task of texture classification is very alike with classical image classification. The only difference might be the data. Some works contribute to describe the feature of some common textures with one or several adjectives as the output of the model to do the supervised classification [29]. Recent studies focused on unsupervised approach to do classification of textures, like [30] proposed deep clustering using metric learning and was used in [31,34] to do the classification of Mars surface texture. Our work of texture extraction refers some methods of texture classification.

## 3. Proposed Method

### 3.1 Method Overview

Our algorithm consists of five modules. Figure 2 provides an overview of the pipeline of our work. The input of our method is a street view image that contains the building we want to process. The first step is a standard semantic segmentation on the street view image facing the building, in order to separate the façade background from the street view. Then, large amount of random regions will be sampled with a certain width on the façade background part segmentated by the previous semantic segmentation module. These will be regarded as candidate representative texture. These candidates will have a dimension reduction with certain model, in our work it's an augmentated auto encoder (section 3.5), into several vectors. Cluster these vectors, we regard the center point of the largest class may be the most representative one. Compute the distance of all candidates to the center point and multiply with a weighted variable (section 3.7), select the closet candidate to the center point as our final result. Detailed explanation of each step will show below.

### 3.2 Problem Definition

More formally, the input is an image consists one façade $F$. We first detect the façade background part in $F$ to get a mask array $M$, where each value in $M$ is true means current pixel is belong to the façade background. Then we do lots of a random crop on $F$ and $M$ get several sub region pairs $\langle F_i, M_i \rangle$, where $i$ represents the $i$th candidate texture and all pixels in the corresponding $M_i$ should be true, otherwise $\langle F_i, M_i \rangle$ would be discarded. Do a dimenstion reduction on $F_i$ and get latent $R_i$. Do a cluster on $R_i$ and select the largest cluster $S$. Compute the distance between each $R_i$ inside the $S$ to the center point of $S$ and get their distances $D_i$. For each $F_i$, compute a value according to the feature of $F_i$ and get $V_i$. Multiply $D_i$ with $V_i$ and get weighted distances $W_i$. Select the $F_i$ that has the minimum $W_i$ as the final output.

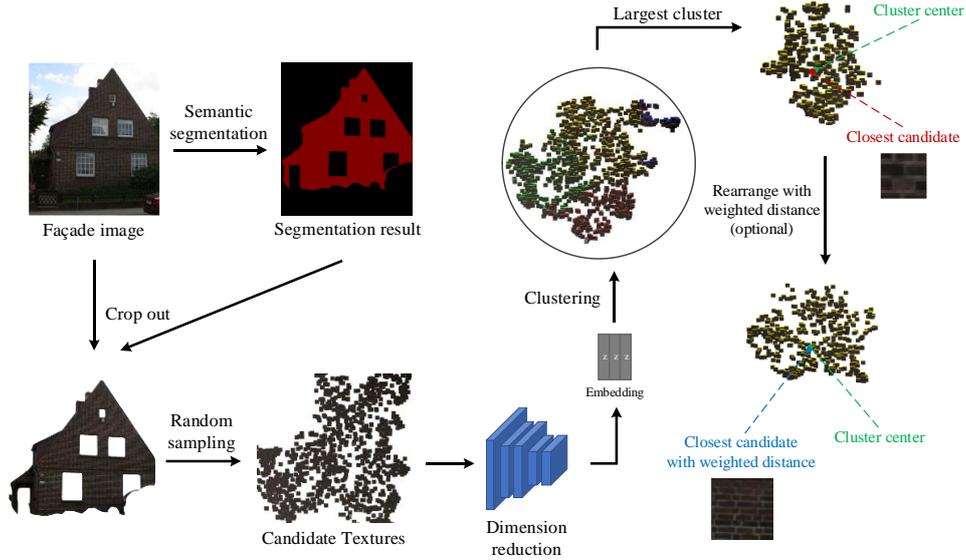

**Figure 2.** Overview of the proposed method pipeline. We input a façade image and output the extracted representative region texture. The pipeline outputs both the texture with and without weighted distance calculation (section 3.7), either one can be used depends on the task.

### 3.3 Step 1 – Façade Parsing with Semantic Segmentation

The façade parsing task is very similar to the task of common object detection or semantic segmentation. For object detection, the target object is every components of the façade and indicates its category. For semantic segmentation, each pixel of the mask image output indicates what kind of component. Considering we need to extract the target texture from exactly the background part, pixel level semantic segmentation meets our need more.

We treat semantic segmentation for façade parsing as an image-to-image mapping problem with the RGB image used as the input and one-hot vectors indicate which category does current pixel belongs to. With the advent of covolutional neural networks, image-to-image translation and prediction problems have become significantly more tractable. Following the success of U-Net [19] which contains an efficient architecture [35], our semantic segmentation module consists of a same pipeline, which the input is the façade image and output is a mask image indicates category of each pixel.

### 3.4 Step 2 – Random Sampling

The representative region texture is inside in the façade background part that filtered out after segmentation. Lots of random sampling are done on the background part. Randomly choose a point inside the façade background part and crop the area centered with this point; more than 90% of pixels inside the area cropped should belong to the façade background part, otherwise the current crop will be discarded and do another crop again.

The width of each crop is between 16~48 pixels. This is determined by the image resolution of the dataset; if a higher resolution dataset is used, the width can be increased as appropriate and vice versa. The random crop process will be repeated around 10000 times per image, and we will get around 10000 crops, which we called as candidate textures. These candidate textures are stored for future processing, and the final result of representative region texture will also be selected from these candidates.

### 3.5 Step 3 – Dimension Reduction with Augmented Pairs

#### 3.5.1 Overview

Dimension reduction techniques reduce the dimensionality of data and removing redundant or irrelevant features. We applied a dimension reduction on candidate textures before clustering. This approach can not only reduce the computational complexity while working as embedding as well. This is based on texture images usually have translation invariance, flip invariance and scale invariance, indicating that if a texture is translated, flipped or scaled, it should be embedded into the similar vector with the original one.

Data augmentation [42, 43] consists of some manipulation of the dataset to create similar data for training in order to increase the amount of data, usually includes spatial transformation such as cropping, flipping, rotation [39] and cutout [41] or appearance transformation such as color disortion and blurring [40,44] in computer vision tasks.

We applied a self-supervised approach [37] that by using some methods from data augmentation including random crop, flip, translate and scale to construct pairs as our training data for the dimension reduction module. Detailed structure and implementation will show below.

### 3.5.2 Augmented Pairs and Augmented Auto Encoder

We construct training pairs with an image and generate its augmented image with scaling and translation, and we called an augmented pair, which is similar to [12, 45] while we did not enable color disortion because we regard the color information that a texture holds is critical for its semantic information which may be destroyed or changed by disortion.

Auto encoder is a commonly used dimension reduction model. We use an auto encoder to train with augmented pairs. In a vanilla auto encoder, the input is the same as the output, while in an augmented auto encoder, the input is the augmented image and the output is the original image. This design will let auto encoder learns from several augmented data and then reconstruct the original one to ensure the model to learn similar representations of augmentation of a image.

### 3.5.3 Network Structure

The size of candidate textures usually don't exceeding 48*48, LeNet [46] already meets our need (if the crop size is larger like 128*128 then backbones like VGG [47] or Xception [48] may be necessary). For encoder part, it's two convolution layers with 16 and 32 as filters. Then flatten the output of convolution layer and feed into a fully-connected layer with size 64 representing the embedding dimension. For decoder part, it's two deconvolution layers with 32 and 16 as filters. After the network is trained, the output of fully-connected layer is the output of dimension reduction.

## 3.6 Step 4 – Clustering

In the real world, the façade usually have various noise. For example, some part of façade may look different from the other due to sun light or rain. In addition, semantic segmentation part may not get an accurate segmentation result. Some pixels that belong to other classes may also be judged as façade background (false positive), which may be intolerable in this task because we wish the extracted texture is not contaminated. Representative region texture should avoid being extracted in these areas.

Therefore, only a random crop on the façade background as the final result may have a high probability of uncorrect or low quality texture because a single random crop may include false positive pixels that doesn't belong to the façade background. Fortunately, in most cases, the number of false positive pixels is usually much less than the whole façade, so most candidate textures generated by random sampling won't contain these false positive pixels. Therefore, we can separate candidate textures with and without false positive pixels by clusterring, and a larger cluster usually has a lower probability of having candidate texture that contains false positive pixels. This is the motivation of clustering.

To be more specific, we cluster the latent outputted by the previous dimension reduction module, and filter out all candidate textures that is not belong to the largest cluster as the amount of false positive pixels is usually much less than the amount of true positive pixels and those true positive already clustered together in the largest class. Subsequent procedures will only be done on candidate textures in the maximum cluster.

We use a standard K-Means as the clustering algorithm. As for the selection of K, we ran clustering with K=3,4,5,6 and select the K that has minimum Davies-Bouldin Index (DBI) [54] for each sample. Candidate textures in the largest cluster are reserved, while others are discarded.

### 3.7 Step 5 – Weighted Distance

#### 3.7.1 Motivation

We evaluate how a candidate texture can represent the façade by measuring the distance to the cluster center, the closer the better. However, this solution has some shortages in pratical tasks. The closest candidate usually has smaller sizes, which may because lower resolution can contain a less specific information and will be clustered closer to the center, but we wish the resolution of candidate texture to be greater if possible in some pratical tasks. Some application will also tile the representative region texture in the process of façade modeling [57] while closest one may not suitable as it may has a lower boundary similarity.

This is the motivation of weighted distance calculation. We do not select candidate whose embedding is closest to the center. The eulicidean distance of each candidate will be multiplied with a weight value and get a weighted distance, and then select the minimum one.

The calculation of weighted distance consists of two parts: width factor and boundary similarity factor. The former will constrain the extracted texture to be larger, and the latter will constrain the texture to have a higher boundary similarity so that to be suitable for tiling. Detailed definition will show below. Note that the weighted distance calculation is an optional step; original distance is enough if the task does not have corresponding requirements.

#### 3.7.2 Definition

Suppose current candidate texture $F_i$ is dimension-reduced to $R_i$ and is clustered and belongs to the largest cluster *S*. Suppose the center point of *S* is *C*. First compute euclidean distance of each $R_i$ to *C*. Then, compute a weight value for each $F_i$. The weight consists of width factor and boundary similarity factor. The width factor weight is defined as follows:

$$T_i = \frac{1}{\sqrt{Width_{F_i} \cdot Height_{F_i}}}. \quad (1)$$

Where $Width_{F_i}$ represents the width of candidate texture $F_i$ and $Height_{F_i}$ represents the height of $F_i$. In our work, the candidate texture is a sqaure, whose width of $F_i$ equals height, so the width factor weight just equals the reciprocal of width. The boundary similarity factor weight is defined as follows:

$$B_i = \frac{2}{\frac{G_i[0,:] \cdot G_i[n-1,:]}{|G_i[0,:]| \cdot |G_i[n-1,:]|} + \frac{G_i[:,0] \cdot G_i[:,m-1]}{|G_i[:,0]| \cdot |G_i[:,m-1]|}}. \quad (2)$$

Where $G_i$ is the grayscale image of $F_i$, *n* represents the height and *m* represents the width. The boundary similarity is the reciprocal of the average of the cosine similarity of up and bottom boundary of grayscale image and left and right boundary. The weight value is calculated as follows:

$$V_i = T_i \cdot B_i. \quad (3)$$

And final weighted distance is then multiplied by the weight value as follows:

$$W_i = D_i \cdot V_i. \quad (4)$$

Where $D_i$ is the original eulicidean distance from each $R_i$ to *C*.

#### 3.7.3 Final Selection

After weighted distance is calculated for each candidate texture, we choose the candidate texture who has the minimum weighted distance as our final result. Output this candidate texture as the result. Some visual results with/without weighted distance will be shown in section 4.

## 4. Experimental Results

### 4.1 Semantic Segmentation for Façade

#### 4.1.1 Data Preparation

For the dataset, we use eTrimsDB [32] and CMP Façade [33] for training. We applied both dataset with two separate semantic segmentation models, where eTrims part is only used in texture extraction because only façade part needs to be detected and CMP part is used in both tasks as it provides a more detailed façade labels.

As for eTrimsDB, we do a data augmentation including random crop, flip and slight color distortion for the eTrims dataset to increase the data distrubution and enhance model generalization ability as training samples are few. Pre-training is also used with cityscapes dataset, while we applied a distribution migration in order to narrow the gap between original dataset (source domain) and final dataset (target domain), which is filtering and crop to preserve more area of the building part in image and reduce the size of roads, people and cars. Each crop should have more than 1/6 size of building and less than 1/3 size of roads.

As for CMP façade, the amount of data is enough for training from scratch so no pre-trained weights are used. CMP façade is an object detection dataset, all of the components labelled are rectangles like bounding boxes rather than pixel-level. Fortunately, latter modules in our work is robust against imprecise segmentation (Section 4.2.4).

#### 4.1.2 Implementation Details

Parameters of our U-Net used doesn't have much differences with the vanilla one, which is two 64, two 128, two 256, two 512 and two 1024 filters in both encoder and decoder while the filter of output layer equals the number of classes. Dropout with 0.5 are used after some of the convolution layers. Adam [36] was used to perform the optimization with learning rate = 0.0001 and decay = 1e-6.

#### 4.1.3 Evaluation Result

We do an evaluation on our semantic segmentation module and compared to related works. Figure 3 shows some visual results. Table 1 shows the quatitive comparsion of segmentation result of façade. Note that we still do semantic segmentation for every class in the dataset for probable future use and comparasion convenience.

Table 1. Comparative results of semantic segmentation for façades. Note that the work of Zhu uses an modified edition of façade dataset while ours uses the vanilla one.

|  | CMP Façade | | eTrimsDB | |
|---|---|---|---|---|
|  | Mean IoU | Pixel-wise Accuracy | Mean IoU | Pixel-wise Accuracy |
| Tylecek et al. [22] | - | 60.3 | - | 82.1 |
| Zhu et al. [53] | **71.6** | - | - | - |
| Ours | 70.4 | **89.4** | **66.4** | **87.3** |

### 4.2 Candidate Selection

#### 4.2.1 Evaluation for Clustering

This section shows evaluation of the clustering process in each subarea of the façade in an overall evaluation for candidate selection. The input to the cluster module is the output from previous dimension reduction module. Figure 4 shows visual results of clustering. The center point of each candidate texture in differenct clusters is marked with different colors. Two t-SNE visualization results are shown in Figure 6.

#### 4.2.2 Comparasion for Weighted Distance

This section shows the comparative results of with and without weighted distance to the center point in final candidate selection. Figure 7 shows some qualitive results. Table 2 shows quatitive comparsion with and without weighted distance.

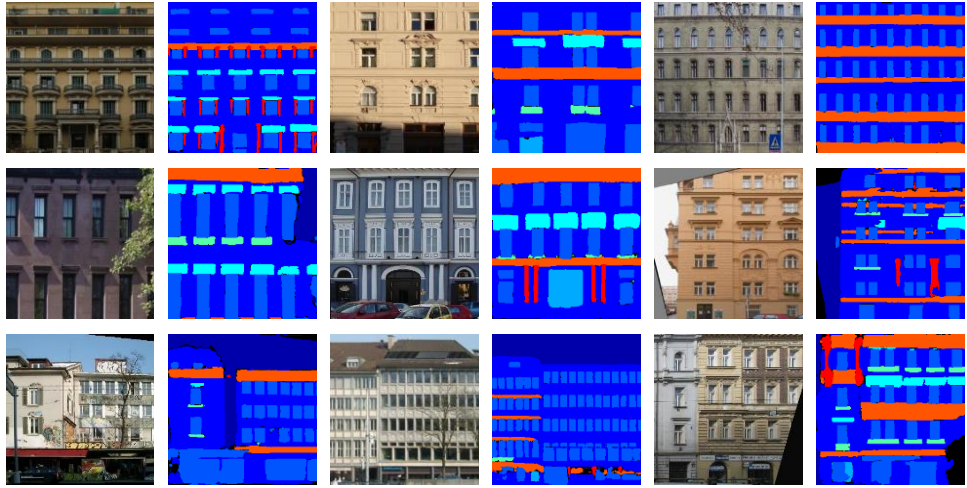

**Figure 3.** Some visual results of semantic segmentation for façade images.

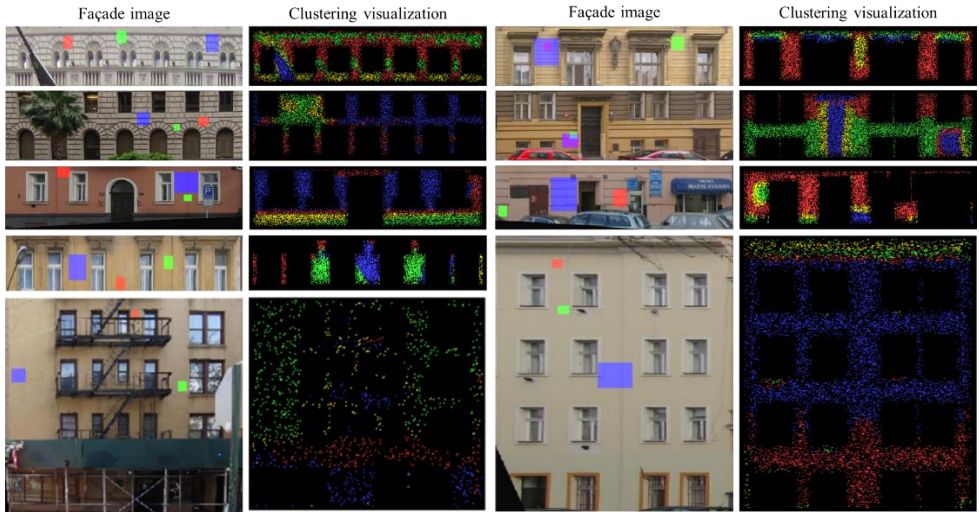

**Figure 4.** Some example experiment results of the clustering module. Left side is the façade image, right side is the visualization result. Each point in the visualization indicates the center point of a sampled candidate texture, candidates in different clusters is marked with different colors. We also mark out the location of extracted texture in the façade image. Red area represents the candidate texture who has the minimum euclidean distance to the center point of largest cluster. Blue area represents the candidate who has the minimum weighted distance to the center point of largest cluster. Green area represents the candidate who has median of distances, which is for reference and is not used in the literature.

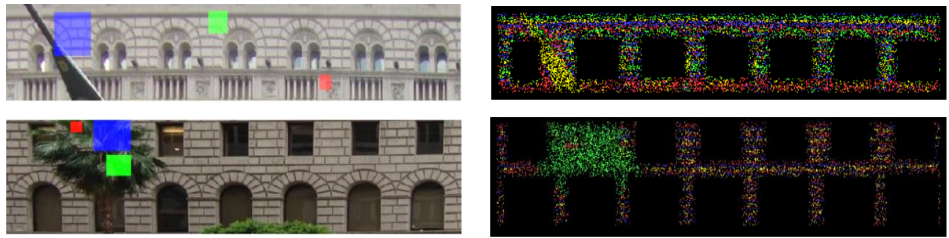

**Figure 5.** Two examples of clustering result of outputs of auto encoder trained without augmented pairs. Bricks in different positions were been clustered to different clusters, since the model cannot recognize the translated image is the same as the original one.

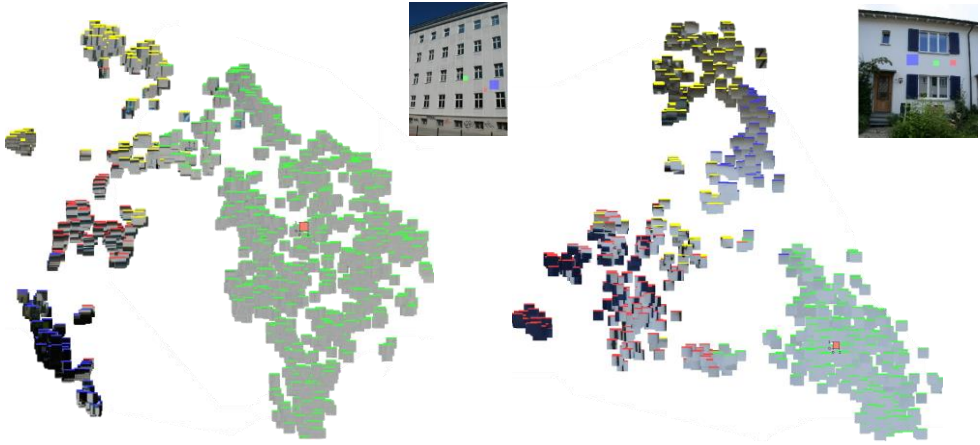

**Figure 6.** t-SNE [52] visualization result of two façade samples. The clusters show good separation, overlap is only existed in candidates who have similar feature. The largest cluster has the most potential representative texture, and we select one who is closest to the center point of the cluster which is colored with salmon in the figure.

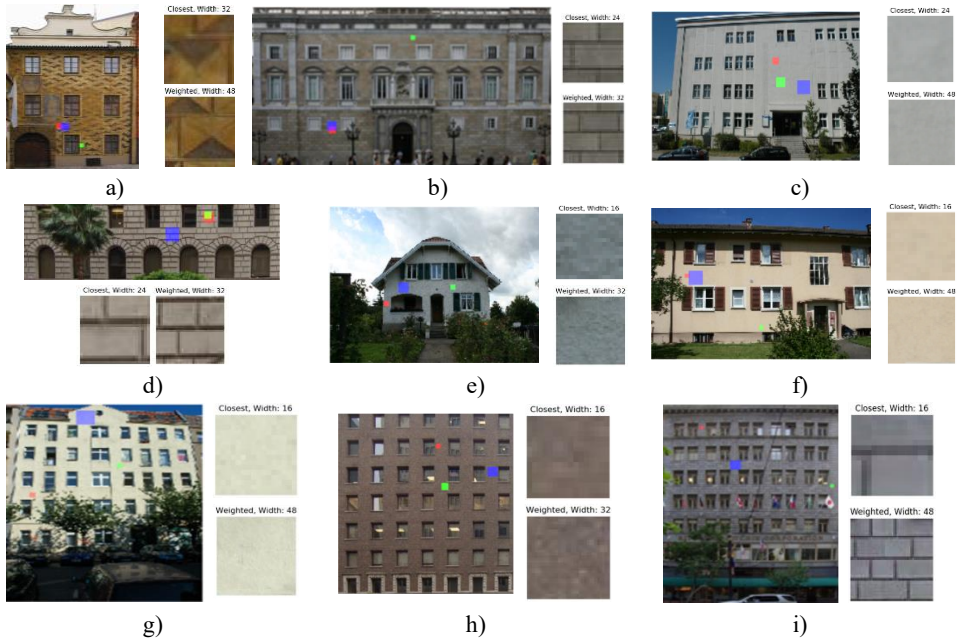

**Figure 7.** Qualitive comparasion of weighted distance. In each sample, left side is the façade image, right side is the extracted texture with (marked as "weighted") and without (marked as "closest") weighted distance. We can see the width of results with weighted distance is usually higher than without one and is more suitable for tiling.

**Table 2.** Qutitive comparsion result with and without weighted distance. Average width of selected candidate increased signifcantly. Average boundary similarity also increased a little.

|  | Selected texture width | Boundary similarity |
| --- | --- | --- |
| w/o weighted distance | 28.67 | 95.8% |
| w weighted distance | 46.53 | 98.0% |

### 4.2.3 Quantitative Evaluation

We do an overall quantitative evaluation with images in CMP and eTrims dataset. Due to the lack of supervised annotations available for our data, we measure the performance of our model by scoring manually and calculate the score proportion of each interval. The score are set from 0 to 10 representing how the texture extracted from the façade image can represent the main texture feature of the façade. Higher scores are better. Detailed scoring rules are as follows:
- 9-10: Both the extracted representative texture with and without weighted distance should well represent the texture feature of original façade. "Well represent" is defined as larger than 80% of the area should belong to the façade part and can reflect the feature of façade.
- 8-9: One of the extracted representative texture with and without weighted distance can well represent the texture feature of original façade, while another can roughly represent the façade.
- 6-8: The extracted representative texture can roughly represent the texture feature of original façade, or one of the extracted texture can well represent while another can not well represent. "Roughly represent" is defined as larger than half of the area of texture can well represent while some part of extracted image cannot represent the façade texture or does not belong to the façade.
- 3-6: The extracted representative texture can not well represent the texture feature of original façade. "Can not well represent" is defined as less than half but more than 25% of the area extracted image can represent the façade texture feature.
- 0-3: The extracted representative texture doesn't represent the texture feature of original façade at all. "Doesn't represent at all" is defined as less than 25% of the extracted area do not represent the façade texture feature or do not belong to the façade.

We also do a comparasion with random cropping on the façade background part. Since a random crop does not have a weighted distance, to be fair for comparasion, we adopted two crops with width 24 and 48 which is close to the average width of texture without/with weighted distance (table 2), and defined scoring rules as follows:
- 9-10: Both the crop with width 24 and 48 in façade should well represent the texture feature of original façade.
- 8-9: One of the crop with width 24 or 48 should well represent the texture feature of façade, while another can roughly represent the façade.
- 6-8: The crop with width 24 and 48 can roughly represent the façade, or one of the crop with width 24 or 48 can well represent while another cannot well represent.
- 3-6: The crop with width 24 and 48 cannot well represent the façade.
- 0-3: The crop with width 24 and 48 cannot represent the façade at all.

Since we don't have actual ground truth or labels available for the retrieved images, we score each output manually accroding to the scoring rules above. Statistical results are shown in Table 3. We can see that our method has a significant improvement compared with extracting texture by just a random crop on façade.

**Table 3.** Quatitive statistical scoring results for overall candidate texture selection process. Score is marked from 0 to 10, the higher the better. Left 5 columns are the proportion of samples in each score range. Right column is the average score.

| Score Range | 9-10 | 8-9 | 6-8 | 3-6 | 0-3 | Avg. Score |
|---|---|---|---|---|---|---|
| **Our method** | 0.58 | 0.14 | 0.12 | 0.06 | 0.09 | **8.1** |
| **Random crop** | 0.17 | 0.09 | 0.34 | 0.21 | 0.19 | 5.6 |

### 4.2.4 Robustness Experiment to Noise

In a real world application, a façade sometimes has noise like sunlight, rain or water pipes on the wall, making some part of façade looks different. Representative textures should avoid being selected in areas containing noises, or specifically, the false positive pixels. Fortunately, the clustering process can handle this as samples not in maxmium cluster are discarded, as we mentioned in section 3.6. To determine how much robustness our method has to the noise, we select 135 images from CMP dataset that have black or gray masks on façade, which is suitable for an anti-noise test. Some visual results are shown in figure 8 and statistical scoring results of these samples are shown in table 4. The result shows that our model can suppress slight noise in image and still gets a correct extracted texture in most cases.

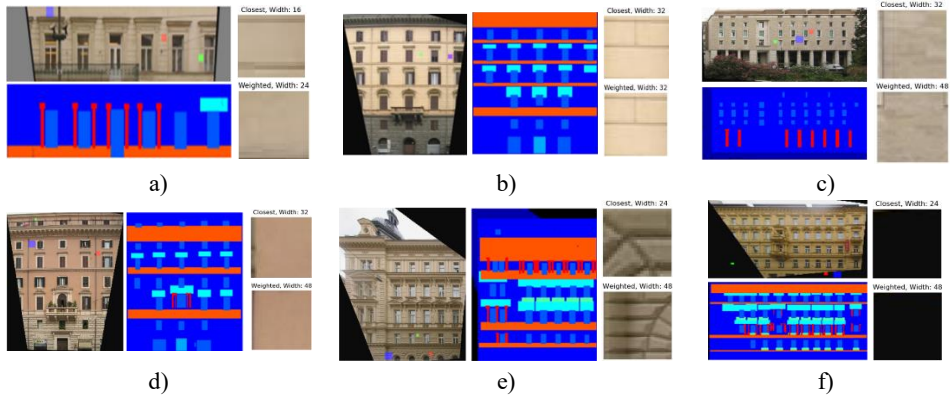

**Figure 8.** Visual results on façade images with noise. We show six examples, in each example left side is the façade image and the semantic information provided, the right side is the extracted texture. The a)-e) are positive results that our model is robust to noise and f) is a negative example that our method fails to extract texture in correct areas.

**Table 4.** Scoring results of 135 images chosen from CMP façade dataset which has noise in façade image.

| Score range | 9-10 | 8-9 | 6-8 | 3-6 | 0-3 | Avg. score |
|---|---|---|---|---|---|---|
| Proportion | 0.51 | 0.15 | 0.11 | 0.08 | 0.12 | 7.8 |

#### 4.2.5 Experiment on General Repetitive Images

The proposed method is also suitable for extracting representative texture for any repetitive images, by removing the semantic segmentation part and input the image directly to random sampling module. The model will look for a maxmium cluster and select a candidate texture closest to the cluster center point which works the same as before. We show some examples in figure 9.

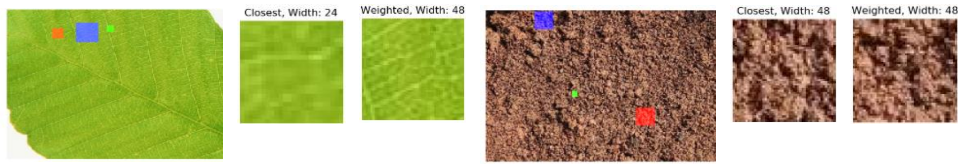

**Figure 9.** Two examples of extracting representative texture from general repetitive images.

### 4.3 Limitaion Analysis

**Multiple Representative Textures.** Our method will regard an input image as one façade and extract one representative texture for this façade. But if the façade has two or more areas that have different representative textures and we wish to extract different textures separately? We noticed that the cornice part in CMP dataset usually can separate areas that has different representative textures, so we process different subareas separated by cornice part. For simplicity we reuse previous segmentation results trained by CMP dataset and applied a breadth-first search to search for seperated subareas.

This solution can solve most multiple representative while still have some counterexamples that in single area has multiple representative textures. We considered a method that regarding the second or third largest cluster may be another representative textures but fails to figure out how to determine the second or third largest cluster belongs to the façade, not noise parts.

**No Representative Textures.** While our method cannot proceed multiple representative textures problem, our method cannot proceed the situation that a façade that does not have a representative texture as well. The representative texture concept is based on that most facades are repetitive and can be mostly represented by a certain region inside of it, while a typical counterexample is that if a façade has a gradient color as its texture, since gradient cannot be represented by a certain region inside of it.

## 5. Discussion

In this section, we will briefly discuss some application scenarios of representative region texture and corresponding related works.

**Façade Modeling.** Considering the resolution of façade images from street view are usually low, directly paste the original façade texture on target reconstructed model may lead to low resolution of model texture. But if switched to use representative texture, using a certain region inside of façade to tiling the whole façade to represent the building, it is possible to improve the visual quality of façade of a building at the expense of loss some façade details. This is an approach of losing details for visual feasibility; the reconstructed building and façade may not exactly the same as the original one because the representative texture is used to replace low-resolution part of orignal façade.

To better illustrate the process of façade modeling with façade semantic information and representative texture, we present an example workflow in figure 10. The general process can be divided into following steps:
1) Façade parsing. This is very similar to the parsing part in representative extraction. Here we only use CMP dataset as it provides a more detailed façade information. Either semantic segmentation or object detection is available here as an approximate bounding region is enough.
2) Find subareas. This part is similar to "multiple representative textures" mentioned in section 4.3 that we find subareas seperated by cornice, and prepare representative region textures for each subarea.
3) Tiling representative region textures on each subarea. First generating a cuboid that represents the façade body. Then tiling the representative textures onto each façade. The size or ratio depends on the actual scene and needs.
4) Make convex on pillars. The pillar part in parsing result can be processed as a vertical convex strip. Considering the texture of pillars may be different from façade, representative textures can also be extracted and tiled on pillar part.
5) Place prefabs on windows and doors. Door and window part will be replaced as prefab prepared. Sizes of prefab placed will be scaled into the size of corresponding bounding box.

**Textured Voxel.** Many works have studied 3D reconstruction with voxel. Inspired by the success of CNN in processing pixels in 2D images, researches also try to use voxels for models to learn a representation of 3D feature. [49] proposed first using 2D CNN to study a latent from original image, then a LSTM [51] to accept multiple images as input, and last a 3D CNN as decoder to reconstruct 3D voxels. [55] proposed a method based upon latent voxel embeddings of an object which can encode appearance and shape information and a neural rendering network to display scene.

However, using voxel as 3D representation format is very memory and computational consuming. From this perception, how to use voxels with low resolution to represent more 3D features of reconstructed objects seems to become increasingly important. Thus a representation approach for 3D buildings with a textured voxel arised. Each voxel can have a certain texture, not only just single color that we see in pixels. We called this a textured voxel. Because the coordinates representing the position of each voxel are integers, triangles where two voxels overlap can be merged without rendering, which is also efficient for rendering engines.

The representative region texture proposed in the literature would be a good choice for the texture of voxels. Because of the condition of low-resolution of a façade, using a certain representative texture is better than directly paste the original façade image onto reconstructed façade since low-resolution voxel may contains uncontinous scene and may be strange visually if directly paste the original image.

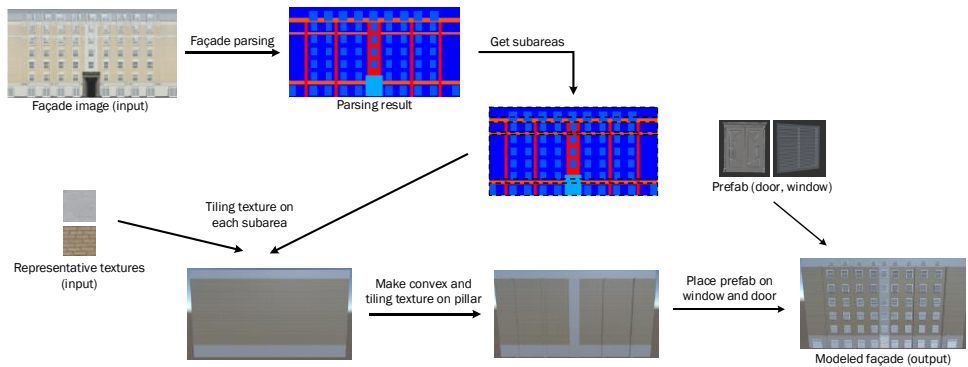

**Figure 10.** The modeling workflow with façade parsing and representative texture. We use an example in the work of Hu et al. [3] that takes from a building in London that has an enhancement of window reconstruction as the façade we proceed. The façade background part is tiled with the representative region texture extracted from the corresponding street view; window and door part is replaced with prefab models; pillar part is processed as convex cuboids. We can see the background part of modeled façade looks more realistic and stereoscopic than directly pasting the image onto the model.

## 6. Conclusion

We have proposed a semi-supervised approach that can extract a representative texture from a façade. The whole process does not require any additional training data; only supervised part is the first semantic segmentation part. Various noise would be very common when dealing with real world façade image and our method has the ability of anti-noise. Our method can also be used to extract representative texture on any repetitive image, not only just façade. However, our method cannot work perfectly in some situations like multiple representative texture extraction without artificial parition. Last we show some examples of using representative texture in practical tasks like façade modeling.

## References

[1] Verdie, Y., Lafarge, F., Alliez, P., 2015. LOD Generation for Urban Scenes. ACM Transactions on Graphics 34, 30:1-30:14.
[2] Fan, H., Wang, Y., Gong, J., 2021. Layout graph model for semantic facade reconstruction using laser point clouds. Geo-spatial Information Science 24, 403-421.
[3] Han Hu, Xinrong Liang, Yulin Ding, Qisen Shang, Bo Xu, Xuming Ge, Min Chen, Ruofei Zhong, Qing Zhu, 2022. Semi-Supervised Adversarial Recognition of Refined Window Structures for Inverse Procedural Façade Modeling. arXiv preprint arXiv:2201.08977 [cs.CV].
[4] Hu, H., Wang, L., Zhang, M., Ding, Y., Zhu, Q., 2020. Fast and Regularized Reconstruction of Building Facades from Street-View Images Using Binary Integer Programming, in: ISPRS Annals of Photogrammetry, Remote Sensing and Spatial Information Sciences, pp. 365-371.
[5] Zolanvari, S.M.I., Laefer, D.F., Natanzi, A.S., 2018. Three-dimensional building façade segmentation and opening area detection from point clouds. ISPRS Journal of Photogrammetry and Remote Sensing 143, 134-149.
[6] Zhu, Q., Shang, Q., Hu, H., Yu, H., Zhong, R., 2021. Structure-aware completion of photogrammetric meshes in urban road environment. ISPRS Journal of Photogrammetry and Remote Sensing 175, 56-70.
[7] Dehbi, Y., Hadiji, F., Groger, G., Kersting, K., Plümer, L., 2017. Statistical Relational Learning of Grammar Rules for 3D Building Reconstruction. Transactions in GIS 21, 134-150.
[8] Wang, R., 2013. 3D building modeling using images and LiDAR: A review. International Journal
of Image and Data Fusion 4, 273-292.
[9] Han, J., Zhu, L., Gao, X., Hu, Z., Zhou, L., Liu, H., Shen, S., 2021. Urban Scene LOD

Vectorized Modeling From Photogrammetry Meshes. IEEE Transactions on Image Processing 30, 7458-7471.

[10] Deng, J., Dong, W., Socher, R., Li, L.J., Li, K., Fei-Fei, L., 2009. ImageNet: A large-scale hierarchical image database, in: 2009 IEEE Conference on Computer Vision and Pattern Recognition, pp. 248-255.

[11] Xie, S., Gu, J., Guo, D., Qi, C.R., Guibas, L., Litany, O., 2020. PointContrast: Unsupervised Pre-training for 3D Point Cloud Understanding, in: Vedaldi, A., Bischof, H., Brox, T., Frahm, J.M. (Eds.), Computer Vision - ECCV 2020, pp. 574-591.

[12] Chen, T., Kornblith, S., Norouzi, M., Hinton, G., 2020. A Simple Framework for Contrastive Learning of Visual Representations, in: Proceedings of the 37th International Conference on Machine Learning (ICML2020), pp. 1597-1607.

[13] Yan. Advantages and disadvantages of end-to-end deep learning and network generalization. https://blog.csdn.net/weixin_32393347/article/details/102515603.

[14] Krizhevsky, A., Sutskever, I., Hinton, G. E., 2012. Imagenet classification with deep convolutional neural networks. Advances in Neural Information Processing Systems, 1097–1105.

[15] Chan, T.-H., Jia, K., Gao, S., Lu, J., Zeng, Z., Ma, Y., 2015. PCANet: A simple deep learning baseline for image classification. IEEE Transactions on Image Processing, 24(12), 5017–5032.

[16] Han, F., Zhu, S.-C., 2008. Bottom-up/top-down image parsing with attribute grammar. IEEE Transactions on Pattern Analysis and Machine Intelligence, 31(1), 59–73.

[17] Ripperda, N., Brenner, C., 2006. Reconstruction of facade structures using a formal grammar and rjmcmc. Joint Pattern Recognition Symposium, Springer, 750–759.

[18] J. Long, E. Shelhamer, and T. Darrell. Fully convolutional networks for semantic segmentation. In CVPR, 2015.

[19] O. Ronneberger, P. Fischer, and T. Brox. U-net: Convolutional networks for biomedical image segmentation. In Proc. Int. Conf. Medical Image Computing and Computer-Assisted Intervention, pages 234–241. Springer, 2015.

[20] L.-C. Chen, G. Papandreou, I. Kokkinos, K. Murphy, and A. L. Yuille. Deeplab: Semantic image segmentation with deep convolutional nets, atrous convolution, and fully connected CRFs. CoRR, abs/1606.00915, 2016.

[21] F. Yu and V. Koltun. Multi-scale context aggregation by dilated convolutions. In ICLR, 2016.

[22] Tylecek, Radim & Sara, Radim. (2013). Spatial Pattern Templates for Recognition of Objects with Regular Structure. 8142. 10.1007/978-3-642-40602-7_39.

[23] Girshick, R., Donahue, J., Darrell, T., Malik, J., 2015. Region-based convolutional networks for accurate object detection and segmentation. IEEE Transactions on Pattern Analysis and Machine Intelligence, 38(1), 142–158.

[24] Alegre, F., Dellaert, F., 2004. A probabilistic approach to the semantic interpretation of building façades.

[25] Girshick, R., Donahue, J., Darrell, T. and Malik, J., 2014. Rich feature hierarchies for accurate object detection and semantic segmentation. In Proceedings of the IEEE conference on computer vision and pattern recognition (pp. 580-587).

[26] Redmon, J., Divvala, S., Girshick, R., Farhadi, A., 2016. You only look once: Unified, real-time object detection. CVPR, 779–788.

[27] He, K., Zhang, X., Ren, S., Sun, J., 2016. Deep residual learning for image recognition. CVPR, 770–778.

[28] Lin, T.-Y., Dollár, P., Girshick, R., He, K., Hariharan, B., Belongie, S., 2017. Feature pyramid networks for object detection. CVPR, 2117–2125.

[29] Hang Zhang, Jia Xue, and Kristin J. Dana. Deep TEN: texture encoding network. CoRR, abs/1612.02844, 2016.

[30] Mathilde Caron, Piotr Bojanowski, Armand Joulin, and Matthijs Douze. Deep clustering for unsupervised learning of visual features. In Proceedings of the European conference on computer vision (ECCV), pages 132–149, 2018.

[31] Tejas Panambur, Deep Chakraborty, Melissa Meyer, Ralph Milliken, Erik Learned-Miller, Mario Parente. Self-Supervised Learning to Guide Scientifically Relevant Categorization of Martian Terrain Images. Earthvision at CVPR Workshops 2022. arXiv:2204.09854.

[32] Korc, F., Förstner, W.: eTRIMS image database for interpreting images of man-made scenes. Tech. Rep. TR-IGG-P-2009-01 (2009).
[33] Tylecek, R.: The CMP facade database. Research Report CTU–CMP–2012–24, Czech Technical University (2012).
[34] Tejas Panambur and Mario Parente. Improved deep clustering of mastcam images using metric learning. In 2021 IEEE International Geoscience and Remote Sensing Symposium IGARSS, pages 2859–2862. IEEE, 2021.
[35] Atapour Abarghouei, Amir & Breckon, Toby. (2018). Real-Time Monocular Depth Estimation Using Synthetic Data with Domain Adaptation via Image Style Transfer. 2800-2810. 10.1109/CVPR.2018.00296.
[36] D. Kingma and J. Ba. Adam: A method for stochastic optimization. In Proc. Int. Conf. Learning Representations, 2014.
[37] Kolesnikov, A., Zhai, X., and Beyer, L. Revisiting self-supervised visual representation learning. In Proceedings of the IEEE conference on Computer Vision and Pattern Recognition, pp. 1920–1929, 2019.
[38] D. Ulyanov, V. Lebedev, A. Vedaldi, and V. S. Lempitsky. Texture networks: Feed-forward synthesis of textures and stylized images. In Proc. Int. Conf. Machine Learning, pages 1349–1357, 2016.
[39] Gidaris, S., Singh, P., and Komodakis, N. Unsupervised representation learning by predicting image rotations. arXiv preprint arXiv:1803.07728, 2018.
[40] Howard, A. G. Some improvements on deep convolutional neural network based image classification. arXiv preprint arXiv:1312.5402, 2013.
[41] DeVries, T. and Taylor, G. W. Improved regularization of convolutional neural networks with cutout. arXiv preprint arXiv:1708.04552, 2017.
[42] Hénaff, O. J., Razavi, A., Doersch, C., Eslami, S., and Oord, A. v. d. Data-efficient image recognition with contrastive predictive coding. arXiv preprint arXiv:1905.09272, 2019.
[43] Bachman, P., Hjelm, R. D., and Buchwalter, W. Learning representations by maximizing mutual information across views. In Advances in Neural Information Processing Systems, pp. 15509–15519, 2019.
[44] Szegedy, C., Liu, W., Jia, Y., Sermanet, P., Reed, S., Anguelov, D., Erhan, D., Vanhoucke, V., and Rabinovich, A. Going deeper with convolutions. In Proceedings of the IEEE conference on computer vision and pattern recognition, pp. 1–9, 2015.
[45] Chen, et al. Big self-supervised models are strong semi-supervised learners. arXiv preprint arXiv:2006.10029 (2020).
[46] Lecun, Yann & Bottou, Leon & Bengio, Y. & Haffner, Patrick. (1998). Gradient-Based Learning Applied to Document Recognition. Proceedings of the IEEE. 86. 2278 - 2324. 10.1109/5.726791.
[47] K. Simonyan and A. Zisserman. Very deep convolutional networks for large-scale image recognition. In ICLR, 2015.
[48] Chollet, Francois. (2017). Xception: Deep Learning with Depthwise Separable Convolutions. 1800-1807. 10.1109/CVPR.2017.195.
[49] Choy, Chris & Xu, Danfei & Gwak, JunYoung & Chen, Kevin & Savarese, Silvio. (2016). 3D-R2N2: A Unified Approach for Single and Multi-view 3D Object Reconstruction. 9912. 628-644. 10.1007/978-3-319-46484-8_38.
[50] Cimpoi, Mircea & Maji, Subhransu & Kokkinos, Iasonas & Mohamed, Sammy & Vedaldi, Andrea. (2013). Describing Textures in the Wild. Proceedings of the IEEE Computer Society Conference on Computer Vision and Pattern Recognition. 10.1109/CVPR.2014.461.
[51] Hochreiter, S., Schmidhuber, J.: Long short-term memory. Neural Comput. 9(8) (November 1997) 1735–1780.
[52] Laurens Van der Maaten and Geoffrey Hinton. Visualizing data using t-sne. Journal of machine learning research, 9(11), 2008.
[53] Zhu, P., Para, W.R., Fruehstueck, A., Femiani, J., Wonka, P., 2020a. Large Scale Architectural Asset Extraction from Panoramic Imagery. IEEE Transactions on Visualization and Computer Graphics doi:10.1109/TVCG.2020.3010694.
[54] Davies, David L., and Donald W. Bouldin. A cluster separation measure. IEEE transactions on pattern analysis and machine intelligence 2 (1979): 224-227.

[55] Tong He, John P. Collomosse, Hailin Jin, Stefano Soatto: DeepVoxels++: Enhancing the Fidelity of Novel View Synthesis from 3D Voxel Embeddings. ACCV (1) 2020: 244-260.

[56] C. A. Vanegas, D. G. Aliaga, and B. Beneš, "Building reconstruction using manhattan-world grammars," in Proc. IEEE Comput. Soc. Conf. Comput. Vis. Pattern Recognit., 2010, pp. 358–365.

[57] Y. I. H. Parish and P. M€uller, "Procedural modeling of cities," Proc. 28th Annu. Conf. Comput. Graph. Interactive Techn., 2001, pp. 301–308.

[58] K. He, G. Gkioxari, P. Dollar, and R. Girshick, "Mask R-CNN," in Proc. IEEE Int. Conf. Comput. Vis., 2017, pp. 2980–2988.